\documentclass[11pt,a4paper]{article}
\usepackage[hyperref]{wmt2022}
\usepackage{times}
\usepackage{latexsym}
\usepackage{bm}
\usepackage{amsmath}
\usepackage{url}
\usepackage{makecell}
\usepackage{multirow}
\usepackage{subfigure}
\usepackage{graphicx} 
\usepackage{ulem}
\usepackage{colortbl}
\usepackage{arydshln}
\usepackage[ruled,linesnumbered]{algorithm2e}

\definecolor{mypink}{rgb}{.99,.91,.95}
\aclfinalcopy
\def\aclpaperid{5}
\usepackage{booktabs}
\usepackage{multirow,multicol}
\DeclareMathOperator*{\argmax}{arg\,max}

\newcommand{\zct}{\color{black}}

\title{\text{\begin{LARGE}{Vega-MT}\end{LARGE}}: The JD Explore Academy Translation System for WMT22}

\author{Changtong Zan$^{\Re,\flat \includegraphics[scale=0.15]{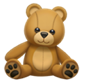}}$,
\ Keqin Peng$^{\sharp  \includegraphics[scale=0.15]{first.png}}$, 
\ Liang Ding$^{\Re  \includegraphics[scale=0.15]{first.png}}$, 
\ Baopu Qiu$^{\natural}$, 
\ Boan Liu$^{\diamondsuit}$, 
\ Shwai He$^{\triangle}$
\\ 
\ \textbf{Qingyu Lu$^{\heartsuit}$, 
\ Zheng Zhang$^{\diamondsuit}$, 
\ Chuang Liu$^{\diamondsuit}$, 
\ Weifeng Liu$^{\flat}$, 
\ Yibing Zhan$^{\Re}$, 
\ Dacheng Tao$^{\Re}$} \\
\ $^{\Re}$JD Explore Academy, JD.com Inc. \\
\ $^{\flat}$China University of Petroleum (East China) 
\ $^{\sharp}$Beihang University
\ $^{\natural}$Nanjing University\\
\ $^{\diamondsuit}$Wuhan University
\ $^{\triangle}$University of Electronic Science and Technology of China
\ $^{\heartsuit}$Southeast University\\
\includegraphics[scale=0.15]{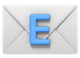} \texttt{dingliang1@jd.com}
}

\begin{document}
\maketitle
\renewcommand{\thefootnote}{\fnsymbol{footnote}}
\footnotetext{\includegraphics[scale=0.15]{first.png} Equal contribution. Work was done when Changtong and Keqin were interning at JD Explore Academy.}
\begin{abstract}
We describe the JD Explore Academy's submission of the WMT 2022 {\zct shared task on general machine translation}. We participated in all high-resource tracks and one medium-resource track, including Chinese$\leftrightarrow$English (Zh$\leftrightarrow$En), German$\leftrightarrow$English (De$\leftrightarrow$En), Czech$\leftrightarrow$English (Cs$\leftrightarrow$En), Russian$\leftrightarrow$English (Ru$\leftrightarrow$En), and Japanese$\leftrightarrow$English (Ja$\leftrightarrow$En). 
\texttt{\bf [Method]}
We push the limit of our previous work -- bidirectional training~\cite{ding2021improving} for translation by scaling up two main factors, \textit{i.e.} language pairs and model sizes, namely the \textbf{Vega-MT} system. As for language pairs, we scale the ``bidirectional’’ up to the ``multidirectional’’ settings, covering all participating languages, to exploit the common knowledge across languages, and transfer them to the downstream bilingual tasks.  As for model sizes, we scale the Transformer-\textsc{BIG} up to the extremely large model that owns nearly 4.7 Billion parameters, to fully enhance the model capacity for our Vega-MT. Also, we adopt the data augmentation strategies, \textit{e.g.} cycle translation~\cite{ding2019usyd} for monolingual data, and bidirectional self-training~\cite{ding2021usyd} for bilingual and monolingual data, to comprehensively exploit the bilingual and monolingual data. To adapt our Vega-MT to the general domain test set, generalization tuning is designed.
\texttt{\bf [Results]}
Based on the official automatic scores\footnote{\url{https://github.com/wmt-conference/wmt22-news-systems/tree/main/scores}} of constrained systems, in terms of the \textbf{\textsc{sacreBLEU}}~\cite{post-2018-call} shown in Figure~\ref{fig:radia}, we got the 1$^{\rm st}$ {\zct place in} \{\begin{small}Zh-En (33.5), En-Zh (49.7), De-En (33.7), En-De (37.8), Cs-En (54.9), En-Cs (41.4) and En-Ru (32.7)\end{small}\}, 2$^{\rm nd}$ {\zct place in} \{\begin{small}Ru-En (45.1) and Ja-En (25.6)\end{small}\}, and 3$^{\rm rd}$ {\zct place in} \{\begin{small}En-Ja(41.5)\end{small}\}, respectively; W.R.T the \textbf{\textsc{COMET}}~\cite{rei-etal-2020-comet}, we got the 1$^{\rm st}$ {\zct place in} \{\begin{small}Zh-En (45.1), En-Zh (61.7), De-En (58.0), En-De (63.2), Cs-En (74.7), Ru-En (64.9), En-Ru (69.6) and En-Ja (65.1)\end{small}\}, 2$^{\rm nd}$ {\zct place in} \{\begin{small}En-Cs (95.3) and Ja-En (40.6)\end{small}\}, respectively. Models will be released to facilitate the MT community through GitHub\footnote{\url{https://github.com/JDEA-NLP/Vega-MT}} and OmniForce Platform\footnote{OmniForce Platform will be launched by JD Explore Academy}.
\end{abstract}

\begin{figure}[htb]
    \centering
    \includegraphics[width=0.43\textwidth]{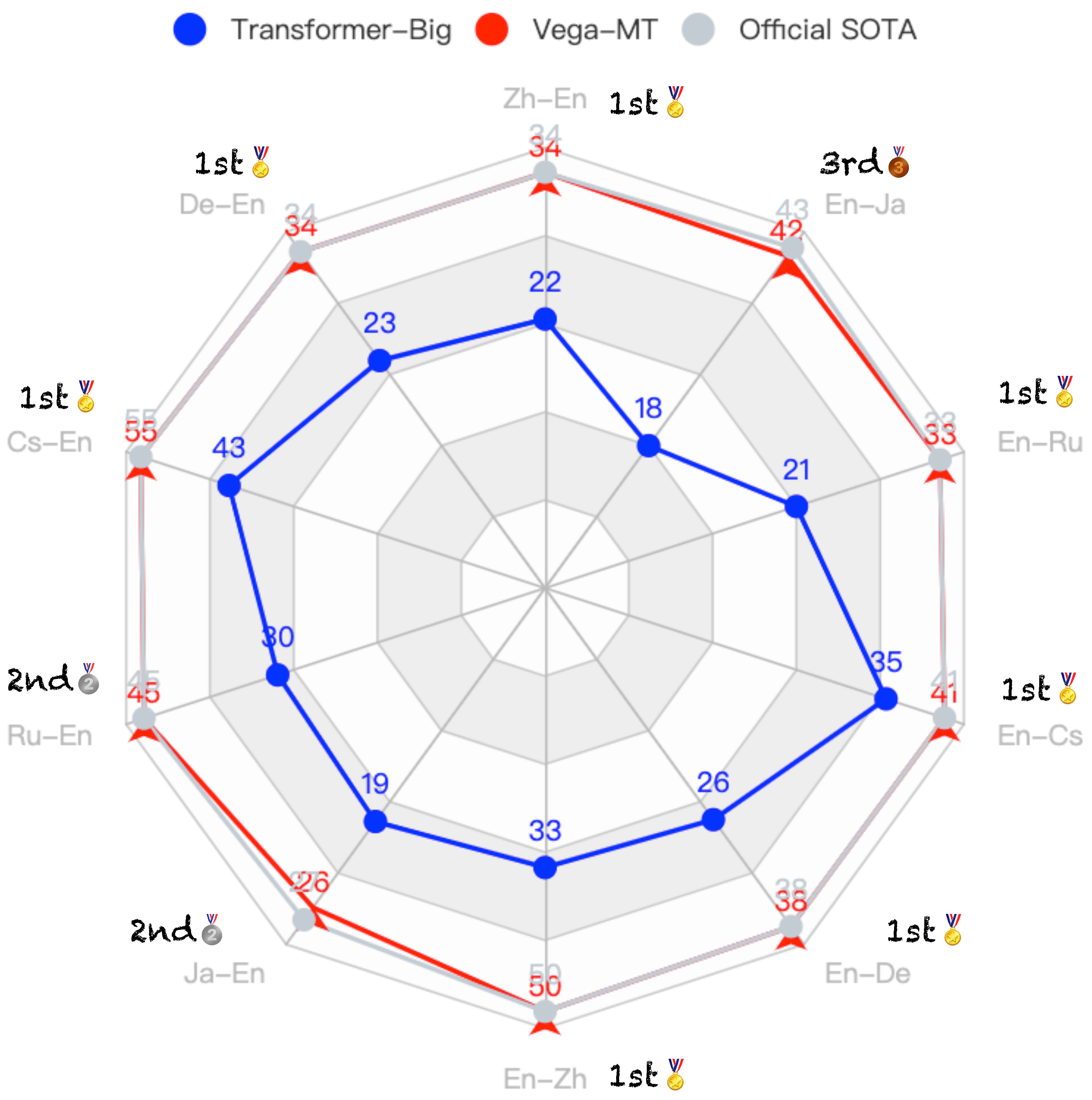}
    \caption{Vega-MT achieves 7 state-of-the-art BLEU points out of 10 high-resource translation tasks among all constrained systems, and significantly outperforms the competitive Transformer-\textsc{BIG} baselines.}
    \label{fig:radia}
\end{figure}

\section{Introduction}
In this year's WMT general translation task, our Vega-MT translation team participated in 10 shared tasks, including Chinese$\leftrightarrow$English (Zh$\leftrightarrow$En), German$\leftrightarrow$English (De$\leftrightarrow$En), Czech$\leftrightarrow$English (Cs$\leftrightarrow$En), Russian$\leftrightarrow$English (Ru$\leftrightarrow$En), and Japanese$\leftrightarrow$English (Ja$\leftrightarrow$En). We use the same model architectures, data strategies and corresponding techniques for all tasks.

We aim to leverage the cross-lingual knowledge through pretraining (PT) to improve the high-resource downstream bilingual tasks. Although recent works~\cite{song2019mass,bart2020,Liu:2020mbart,wang-etal-2022-understanding} attempt to leverage sequence-to-sequence PT for neural machine translation (\citealp[NMT;][]{bahdanau2014neural,DBLP:journals/corr/GehringAGYD17,DBLP:journals/corr/VaswaniSPUJGKP17}) by using a large amount of unlabeled (\textit{i.e.} monolingual) data, \citet{Zan2022PTvsRI} show that it usually fails to achieve notable gains (sometimes, even worse) on resource-rich NMT on par with their random-initialization counterpart, which is consistent with our preliminary experiments. \citet{ding2021improving} show that bidirectional pretrained model as initialization for downstream bilingual tasks could consistently achieve significantly better performance. It is natural to assume that scaling the ``bidirectional'' to the ``multidirectional'' setting with \{1) \textit{multilingual pretraining} and 2) \textit{large enough model capacity}\} could benefit the downstream resource-rich bilingual translations. \citet{facebook2021} and \citet{lin2020pre} also provide empirical evidences to support our motivation of supervised multilingual pretraining. Different from \citet{facebook2021} that explores the effectiveness of multilingual training, we show that further tuning on the bilingual downstream task provide more in-domain knowledge and thus could gain better translation quality. Compared with \citet{lin2020pre}, our model do not require any alignment information during pretraining, which will consume more extra time and computation resources, making our strategy {\zct flexible to be applied to any language}.

For model frameworks in \S\ref{ssec:nmt}, we tried autoregressive neural machine translation, including Transformer-\textsc{BIG} and -\textsc{XL}~\cite{transformer}, and non-autoregressive translation models~\cite{gu2018non}, where the Transformer-\textsc{XL} is employed as the foundation model and autoregressive \textsc{Big} and {\zct non-autoregressive} models are used during augmenting. 
For the core training strategy of our Vega-MT, we cast the multilingual pretraining as foundation models in \S\ref{ssec:foundationmodel}, including \textsc{Multi-Directional Pretraining} (\S\ref{ssec:multi-pt}) and \textsc{Specific-Directional Finetuning} (\S\ref{ssec:sd-ft}). 
For data augmentation strategies, we employ \textsc{Cycle Translation} (\S\ref{ssec:ct}) and \textsc{Bidirectional Self-Training} (\S\ref{ssec:bi-selftraining}) for both monolingual and parallel data.
In order to adapt our Vega-MT to the general domains, we employ \textsc{Greedy Based Ensembling} (\S\ref{ssec:ensemble}), \textsc{Generalization Finetuning} (\S\ref{ssec:general-ft}) and \textsc{Post-processing} (\S\ref{ssec:post}) strategies.

The subsequent paper is designed as follows. We introduce the major approaches we used in Section~\ref{sec:app}. In Section~\ref{sec:data}, we provide the data description. We also present the  experimental settings and results in Section~\ref{sec:exp}. Conclusions are described in Section~\ref{sec:con}.

\begin{figure*}[htb]
    \centering
    \includegraphics[width=0.98\textwidth]{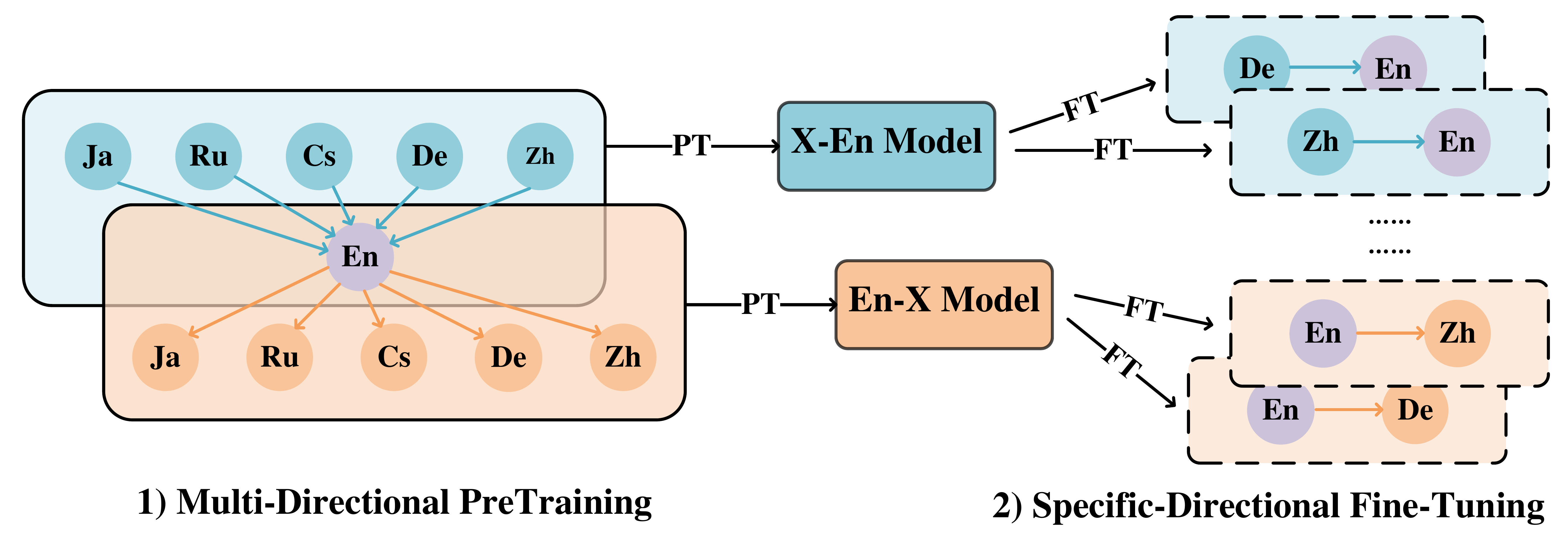}
    \caption{The schematic structure of the two main stages of the Vega-MT. }
    \label{fig:system}
\end{figure*}

\section{Approaches}
\label{sec:app}

\subsection{Neural Machine Translation Frameworks}
\label{ssec:nmt}
The neural machine translation task aims to transform a source language sentence into the target language with a neural network. There are several generation paradigms for translation, \textit{e.g.} Autoregressive Translation (\citealp[AT,][]{rnnsearch,transformer}) and Non-Autoregressive Translation (\citealp[NAT,][]{gu2018non}).

\begin{table}[]
    \centering
    \begin{tabular}{lccc}
    \toprule
    \textbf{} & \bf $\mathcal{M}_{\textbf{Base}}$ & \bf $\mathcal{M}_{\textbf{Big}}$ & \bf $\mathcal{M}_{\textbf{XL}}$  \\ \hline
    {\bf $\#$Stack}  &     6    & 6     & 24    \\
    {\bf $\#$Hidden\_Size}  &    512       & 1024  & 2048  \\
    {\bf $\#$FFN\_Size}  &     2048        & 4096  & 16384 \\
    {\bf $\#$Heads}  &    8    & 16    & 32    \\
    \bottomrule
    \end{tabular}
    \caption{\label{tab:models} Model differences among base ( $\mathcal{M}_{\textbf{Base}}$ ), big ($\mathcal{M}_{\textbf{Big}}$) and extremely large ($\mathcal{M}_{\textbf{XL}}$).}
\end{table}

\paragraph{Autoregressive Translation}
Given a source sentence $\bf x$, an NMT model generates each target word ${\bf y}_t$ conditioned on previously generated ones ${\bf y}_{<t}$. Accordingly, the probability of generating $\bf y$ is computed as:
\begin{equation}
    p({\bf y}|{\bf x})
    =\prod_{t=1}^{T}p({\bf y}_t|{\bf x},{\bf y}_{<t}; \theta)
    \label{eq:standard}
\end{equation}
where $T$ is the length of the target sequence and the parameters $\theta$ are trained to maximize the likelihood of a set of training examples according to $\mathcal{L}(\theta) = \argmax_{\theta} \log p({\bf y}|{\bf x}; \theta)$. Typically, we choose Transformer~\cite{transformer} as its state-of-the-art performance and scalability. We carefully employ the standard Transformer-\textsc{BASE} ($\mathcal{M}_{\textbf{Base}}$) and Transformer-\textsc{BIG} ($\mathcal{M}_{\textbf{Big}}$) in the preliminary studies, and also scale the framework up to an extremely large setting~\cite{facebook2021} -- Transformer-\textsc{XL} ($\mathcal{M}_{\textbf{XL}}$) to maintain powerful model capacity {\zct (see Table~\ref{tab:models}) }.
In Vega-MT, we utilized {\zct the} autoregressive translation (AT) model with $\mathcal{M}_{\textbf{Big}}$ and $\mathcal{M}_{\textbf{XL}}$ for {multi-directional pretraining} (\S\ref{ssec:multi-pt}), {specific-directional finetuning} (\S\ref{ssec:sd-ft}), {bidirectional self-training} (\S\ref{ssec:bi-selftraining}) and {generalization fine-tuning} (\S\ref{ssec:general-ft}) as its powerful modelling ability and generation accuracy.

\paragraph{Non-Autoregressive Translation}
Different to autoregressive translation~\cite[AT]{rnnsearch,transformer} models that generate each target word conditioned on previously generated ones, non-autoregressive translation~\cite[NAT]{gu2018non} models break the autoregressive factorization and produce the target words in parallel.
Given a source sentence $\bf x$, the probability of generating its target sentence $\bf y$ with length $T$ is defined by NAT as:
\begin{equation}
    p({\bf y}|{\bf x})=p_L(T|{\bf x}; \theta) \prod_{t=1}^{T}p({\bf y}_t|{\bf x}; \theta)
\end{equation}
where $p_L(\cdot)$ is a separate conditional distribution to predict the length of target sequence. 
Typicallly, most NAT models are implemented upon the framework of $\mathcal{M}_{\textbf{Base}}$.
We utilized {\zct the} NAT for {bidirectional self-training} (\S\ref{ssec:bi-selftraining}) as NAT can nicely avoid the error accumulation problems during generation, and generate diverse synthetic samples. 
Also, we employ several advanced structure~\cite{gu2019levenshtein,ding-etal-2020-context} (\textit{Levenshtein} with source local context modelling) and advanced training strategies~\cite{Ding2020Progressive,Ding2020Rejuvenating,Ding2020UnderstandingAI,ding2022redistributing,ding2022neural} to obtain high quality and diverse translations.

\subsection{Multidirectional Pretraining as Foundation Models}
\label{ssec:foundationmodel}

This section illustrates how we scale the ``bidirectional'' training in \citet{ding2021improving} up to ``multidirectional'' pretraining with all high-resource parallel corpora, including Zh, De, Cs, Ru, Ja to/from En. The pretrained foundation models will be {\zct finetuned for} the downstream specific-directional task, \textit{e.g.} Zh-En. Such two-stage scheme is shown in Figure~\ref{fig:system}.

\subsubsection{Multi-Directional Pretraining}
\label{ssec:multi-pt}
Recent works on real-world \texttt{WMT} translation datasets have verified that {\zct it is possible to transfer} the pretrained cross-lingual knowledge to the downstream tasks with {\zct the} pretrain-finetune paradigm, hence improving performance and generalization ability~\cite{ding2022redistributing, ding2022improving, wang-etal-2020-tencent}.

Here, we propose multi-directional pretraining by extending Bidirectional Pretraining \cite[BiT]{ding2021improving} to utilize multiple translation {\zct corpora} of different languages. Compared with BiT, multi-directional pretraining could utilize the cross-lingual knowledge among more languages, thus further exploiting the cross-language knowledge and facilitating the downstream transferring.
The main modifications could be summarized {\zct twofold}: 

1) We increase language numbers to utilize the cross-lingual knowledge of various languages. 
The straight setting for multi-directional pretraining is multi-lingual translation, which is divided into Many-to-Many (M2M), One-to-Many (O2M), and Many-to-One (M2O), according to the language number that the model supports. M2M has potential of capturing more cross-lingual knowledge from $N*N$ pairs compared with $N*1/1*N$ pairs of M2O/O2M but usually leads to worse performance because of the imbalanced language data distribution question~\cite{freitag2020complete}.
Inspired by~\cite{facebook2021}, we focus on pretraining two separate systems, including English-to-Many and Many-to-English. 
We also prepend the corresponding language token to source \& target sentences.

2) We further expand model size to an extremely large setting. While enjoying the benefit of cross-lingual knowledge transferring, the difficulty of modeling extremely large-scale data and language-specific feature pushes us to enlarge Transformer-\textsc{BIG} to an extremely large size {\zct (4.7 Billion parameters, see Table~\ref{tab:models})}. 
This ensures our models are capable of better mastering multiple translation corpus.

\subsubsection{Specific-Directional Finetuning}
\label{ssec:sd-ft}
{\zct The off-target problem, which widely exits in multilingual translation systems~\cite{yang-etal-2021-improving-multilingual}, indicates model often generates the translation with some non-target words.}
To reduce non-target word translation ratio in multi-directional pretrained models, we consider a two-stage specific-directional finetuning strategy. As shown in Figure~\ref{fig:system}, the English source/target model is tuned with an English source/target bilingual corpus.

Specifically, we first replace the multilingual embedding with a bilingual one. To fit model and bilingual vocabulary, we freeze all parameters of the Transformer backbone and only tune embedding layers in this stage. Next, we employ full model finetuning on large-scale translation corpus. This allows {\zct the} model to fully adapt to the specific directional translation task, thus further achieving gains. To balance both finetune stages, we set the ratios of update step as $1:4$ for embedding- and full model-tuning, respectively.

For future work during specific directional finetuning, it will be interesting to design tuning data order~\cite{liu-etal-2020-norm,zhou-etal-2021-self} by leveraging the learning difficulty of each training sample estimated in the pretraining stage.

\subsection{Data Augmentation Strategies}
\label{ssec:DA}
In Vega-MT, we consider augmenting both the parallel and monolingual data comprehensively. Specifically, we employ the cycle translation~\cite{ding2019usyd} for regenerating the low-quality \textit{monolingual data}, and adopt bidirectional self-training~\cite{ding2021usyd} to distill, diversify \textit{both the monolingual and parallel data}.

\subsubsection{Cycle Translation for Mono. Data}
\label{ssec:ct}
\begin{figure}[t!]
    \centering
    \includegraphics[width=0.45\textwidth]{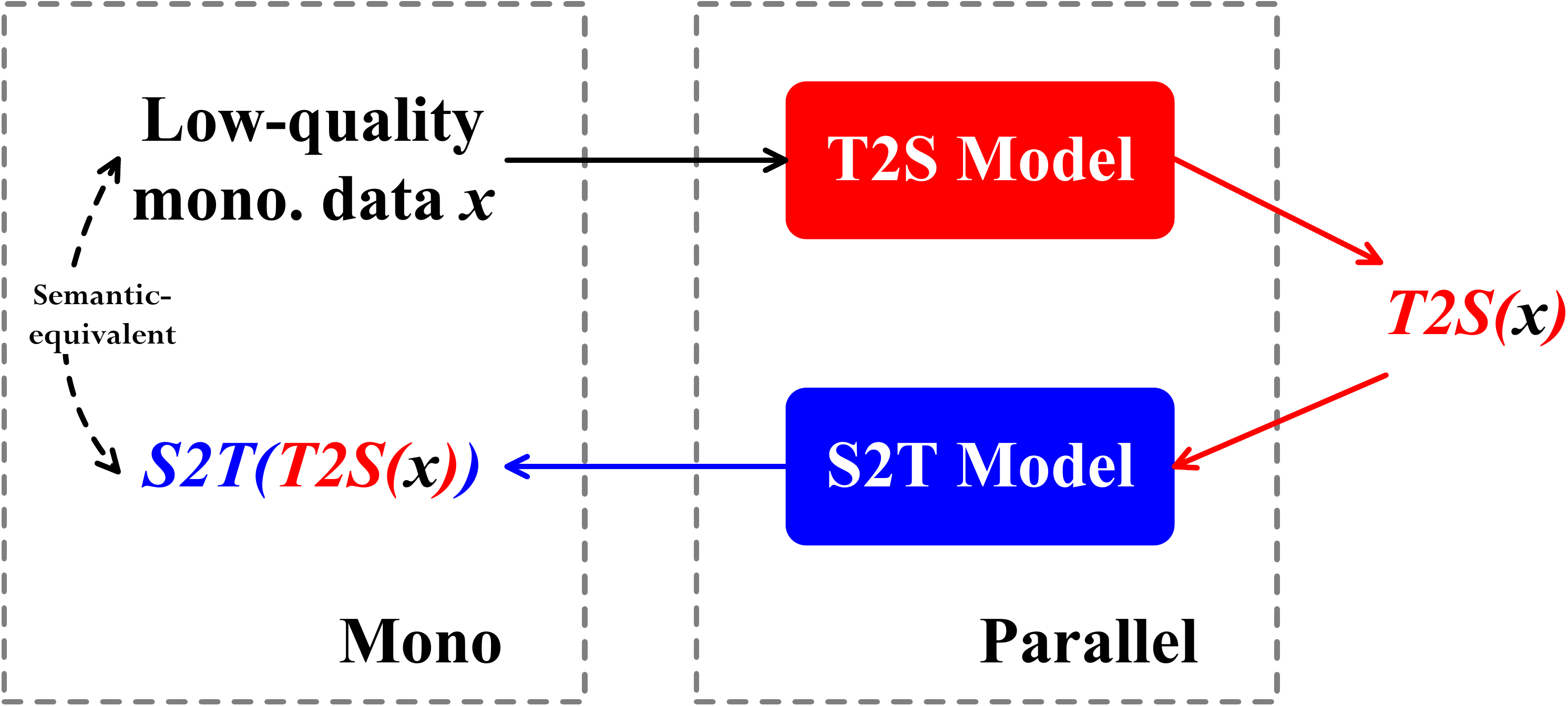}
    \caption{The Cycle Translation process, into which we feed the low quality monolingual data $x$, and then correspondingly obtain the improved data $\mathcal{C}\mathcal{T}(x)$ (denoted as $S2T(T2S(x))$). Note that models marked in \textcolor{red}{red} and \textcolor{blue}{blue} represent the target-to-source and source-to-target model trained with $\mathcal{M}_\textbf{Big}$. The dotted double-headed arrow between the input $\mathbf{x}$ and the final output $\mathcal{C}\mathcal{T}(x)$ means they share the semantic but differ in fluency.}
    \label{fig:cycle-translation}
\end{figure}

There is a large amount of monolingual data incomplete or grammatically incorrect. To fully leverage such part of monolingual data for better data augmentation, \textit{e.g.} back translation~\cite{sennrich-etal-2016-improving} or {\zct sequence -level} knowledge distillation~\cite{kim-rush-2016-sequence}, we adopt Cycle Translation~\cite{ding2019usyd} (denoted as $\mathcal{C}\mathcal{T}(\cdot)$, as Figure~\ref{fig:cycle-translation}) to improve the monolingual data below the quality-threshold (the latter 50\% will be cycle translated according to \citet{ding2019usyd}'s optimal setting). We give an example in Table~\ref{tab:sentences} to clearly show how the cycle translation improves the quality of the sentence.

\begin{table}[t!]
    \begin{center}
    \begin{tabular}{lp{6.4cm}}
    \toprule
    \# & \textbf{Cycle Translated Sentence ``1''$\rightarrow$``2''}\\
    \midrule
    1 & \textit{She stuck to her principles even when some suggest that in an environment often considered devoid of such thing there are little point.}\\ 
    \midrule
    2 & \textit{She insists on her own principles, even if some people think that it doesn't make sense in an environment that is often considered to be absent.}\\ 
    \bottomrule
    \end{tabular}
    \end{center}
    \caption{\label{tab:sentences}Example of difference between original sentence (line 1) and cycle translated result (line 2). Pre-trained BERT model using all available English corpora show that the $\mathcal{L}oss$ decreased from 6.98 to 1.52.}
    \end{table}

\subsubsection{Bidirectional Self-Training for Both Mono\&Para Data}
\label{ssec:bi-selftraining}
Currently, data-level methods have attracted the attention of the community, including exploiting the parallel and monolingual data.
The most representative approaches include: 
{\zct
\begin{itemize}
    \item Back Translation (\textbf{BT},~\citealt{sennrich-etal-2016-improving}) introduces the target-side monolingual data by translating with an inverse translation model, and combines the synthetic data with parallel data;
    \item Knowledge Distillation (\textbf{KD},~\citealt{kim-rush-2016-sequence}) generates the synthetic data with sequence-level knowledge distillation; 
    \item Data Diversification (\textbf{DD},~\citealt{NEURIPS2020_7221e5c8}) diversifies the data by applying KD and BT on parallel data. 
\end{itemize}
}
Clearly, self-training is at the core of above approaches, that is, they generate the synthetic data either from source to target or reversely, with either monolingual or bilingual data. 

To this end, we employ the bidirectional self-training~\cite{ding2021usyd,liao-etal-2020-multi} strategy for both parallel and monolingual data (including source and target, respectively).
Specifically, baseline AT models with $\mathcal{M}_{\textbf{Big}}$ setting and NAT models with $\mathcal{M}_{\textbf{Base}}$ setting are trained with original (distilled for NAT) parallel data in the first iteration, and based on these forward- and backward-teachers, all available source \& target language sentences can be used to generate the corresponding synthetic target \& source sentences. The authentic and synthetic data (generated by AT and NAT models) are then concatenated to train the second round AT and NAT models. 
We run the bidirectional self-training by totally 2 rounds for each translation direction. And for each round, we train 3 forward- and 3 backward- AT models, and 1 forward- and backward- NAT models to perform self-training. 
In this way, the amount of bidirectional synthetic data will be 8x larger than the original parallel and monolingual data.

\begin{algorithm}[t]
\KwIn{Single Model $M_n$, ~~~~~~~~~~~~~~~~General Seed $D$=\{$D_1$, $D_2$..$D_k$\}, ~~~~~~~~~~~~~~Ensemble $N$ models $E_N$.}
\KwOut{New Model $M_{n}^{'}$}
        $t:=0$\\
        \While {not convergence}{
      Translate $D_1$ with $E_N$ and get $D_{1}^{E_N}$\\
      ..\\
      Translate $D_k$ with $E_N$ and get $D_{k}^{E_N}$\\
      $D^{E_N} = D_{1}^{E_N} \cup .. D_{k}^{E_N}$\\
      Train $M_n$ on $D \cup D^{E_N}$ and get $M_{n}^{'}$, then $M_n=M_{n}^{'}$ \\ 
        $t := t + 1$\\      
        }
\caption{Generalization Finetuning {\zct with} Iteratively Transductive Ensemble} 
\label{alg:1}
\end{algorithm}

\subsection{Generalization Adaptation for Downstream Translation}
To adapt Vega-MT to the general domain translation task, we employ several strategies, including ensembling, generalization finetuning, and post-processing. Note that in our preliminary study, we find that noisy channel reranking with the target-to-source MT model and language model does not work in our setting, thus we have not reranked the results in the final submission.

\subsubsection{Greedy Based Ensembling}
\label{ssec:ensemble}
Greedy based ensembling adopts an easy operable greedy-base strategy to search for a better single model combinations on the development set, which consistently shows better performance than simply average in our preliminary study, therefore we technically follow the instruction of~\citet{deng-etal-2018-alibabas} to choose the optimal combination of checkpoints to enhance the generalization and boost performance of the final model. We refer to this method as ``Ensemble'' in the following.

\begin{table}[t!]
    \begin{center}
    \begin{tabular}{lp{5.74cm}}
    \toprule
    {\sc \textbf{src}} & \textit{Siltalan edellinen kausi liigassa oli \uwave{2006-07}}\\
    \midrule
    {\sc \textbf{hyp}} & \textit{Siltala's previous season in the league was \uwave{2006 at 07}}\\ 
    \midrule
    ~\textbf{+post} & \textit{Siltala's previous season in the league was \uwave{2006-07}}\\ 
    \bottomrule
    \end{tabular}
    \end{center}
    \caption{\label{tab:post-process}Example of the effectiveness of post-processing in handling inconsistent number translation.}
\end{table}

\begin{table}[t!] 
    \centering
    \begin{tabular}{lrr} 
    \toprule 
    \textbf{Languages} & \textbf{\# Sents} & \textbf{\# Ave. Len.}\\ 
    \midrule 
    \multicolumn{3}{c}{\textit{Parallel}} \\ 
    \midrule 
    \textsc{Zh-En} & ~46,590,547 & 22.8/27.1 \\ 
    \textsc{De-En} & 292,020,383 & 22.9/21.7 \\ 
    \textsc{Cs-En} & ~88,244,832 & 20.5/19.9 \\ 
    \textsc{Ru-En} & ~98,454,430 & 28.5/27.8 \\ 
    \textsc{Ja-En} & ~28,943,024 & 26.2/28.0  \\ 
    \midrule 
    \multicolumn{3}{c}{\textit{Monolingual}} \\ 
    \midrule 
    \textsc{En} & 1,384,791,758 & 21.3 \\ 
    \textsc{Zh} & 1,346,538,572 & 25.8 \\ 
    \textsc{De} & 5,612,161,001 & 23.2 \\ 
    \textsc{Cs} & 444,049,843   & 19.7 \\ 
    \textsc{Ru} & 8,351,860,471 & 28.5 \\ 
    \textsc{Ja} & 5,534,872,418 & 27.9 \\ 
    \bottomrule 
    \end{tabular}
    \caption{\label{tab:data_statics}Data statistics after pre-processing.}
\end{table}

\begin{table*}[t!]
    \centering 
    \scalebox{1}{
    \begin{tabular}{lcclccl}
    \toprule
    & \multicolumn{3}{c}{\textbf{Zh-En}}  & \multicolumn{3}{c}{\textbf{En-Zh}}  \\ \hline
    \textbf{Models}  & \textbf{W21 test} & \textbf{W22 test} & \textbf{$\Delta$} & \textbf{W21 test} & \textbf{W22 test} & \textbf{$\Delta$} \\ \hline
    \textbf{Transformer-\textsc{BIG} w/ Para.} & 25.3 & 21.9 & - & 25.9 & 33.2 & - \\ \hline
    \textbf{Multi-Directional PT}     & 28.4 & 25.1 &\it +3.2 & 27.1 & 35.7 &\it  +1.9\\
    ~~~{+Specific-Directional FT}     & 29.5 & 26.7 &\it +4.3 & 27.4 & 36.6 &\it  +3.6  \\\hdashline
    ~~~{+Bidirect. Self-Training}     & 30.8 & 29.0 &\it +6.3 & 29.7 & 40.7 &\it  +5.7  \\\hdashline
    ~~~{+Ensemble}                    &\bf 31.1 & 29.8 &\it +6.7 & 30.4 & 41.3 &\it  +6.4  \\
    ~~~{+Generalization FT}           & 30.3 &\bf 33.5 &\it +8.3 & 30.6 & 44.1 &\it +9.0 \\
    ~~~{+Post-Processing}             & 30.5 &\bf 33.5 &\it\textbf{+8.4}&\bf 33.6 &\bf 49.7 &\it \textbf{+13.3} \\
    \bottomrule
    \end{tabular}}
    \caption{\label{tab:ablation} \textbf{Ablation studies of each component on Zh$\leftrightarrow$En} general translation task in terms of SacreBLEU.{\zct We select Transformer-\textsc{BIG} only trained with official parallel data as the baseline.}}
\end{table*}

\begin{table*}[t!]
    \centering 
    \begin{tabular}{lcccccc}
    \toprule 
    \textbf{Models}        & \textbf{Zh$\rightarrow$En} & \textbf{De$\rightarrow$En} & \textbf{Cs$\rightarrow$En} & \textbf{Ru$\rightarrow$En} & \textbf{Ja$\rightarrow$En} & \textbf{$\Delta$} \\\hline
    \textbf{Baseline}      & 21.9            & 23.0            & 42.5            & 30.2            & 19.0     & -\\
    \textbf{Vega-MT}    & \textbf{33.5}   & \textbf{33.7}   & \textbf{54.9}   & 45.1            & 25.6        &\textit{\textbf{+11.2}}\\\hdashline
    {Best Official} & 33.5            & 33.7            & 54.9            &\bf 45.1            &\bf 26.6     & \\\midrule
    \textbf{Models}        & \textbf{En$\rightarrow$Zh} & \textbf{En$\rightarrow$De} & \textbf{En$\rightarrow$Cs} & \textbf{En$\rightarrow$Ru} & \textbf{En$\rightarrow$Ja} & \textbf{$\Delta$}\\\hline
    \textbf{Baseline}      & 33.2            & 26.4            & 34.8            & 20.8            & 17.9     & -\\
    \textbf{Vega-MT}    & \textbf{49.7}   & \textbf{37.8}   & \textbf{41.4}   & \textbf{32.7}   & 41.5        &\textit{\textbf{+14.0}}\\\hdashline
    {Best Official} & 49.7            & 37.8            & 41.4            & 32.7            &\bf 42.5     & \\
    \bottomrule
    \end{tabular}
    \caption{\label{tab:sacrebleu} \textbf{SacreBLEU-Scores of our submissions in WMT2022 general translation task.} ``Baseline'' indicates the performance of the baseline systems. And ``Best Official'' denotes the best results {\zct of constrained systems} in each direction.}
\end{table*}

\subsubsection{Generalization Finetuning}
\label{ssec:general-ft}
As the general domain evaluation is on multi-domain directions, \textit{i.e.} containing (up to) four different domains, we design generalization finetuning strategy to transductively finetune~\cite{Wang2020TransductiveEL} on each domain, and ensemble them into one single model, to empower the general translation ability. 
The proposed generalization finetuning is shown in Algorithm~\ref{alg:1}.
The main difference from Multi-Model \& Multi-Iteration Transductive Ensemble~\cite{wang-etal-2021-tencent} is that the $k_{th}$ domain seed $D_k$ is extracted from the test set using heuristic artificial knowledge. 

\subsubsection{Post-Processing}
\label{ssec:post}
In addition to general post-processing strategies (\textit{e.g.} de-BPE), we also employ a post-processing algorithm~\cite{wang-etal-2018-niutrans} for inconsistent number, date translation, for example, ``\textit{2006-07}'' might be translated to the wrong translation ``\textit{2006 at 07}''. Our post-processing algorithm will search for the best matching number string from the source sentence to replace these types of errors {\zct (see Table~\ref{tab:post-process})}.
Besides, we also conduct punctuation conversion, including convert quotation marks to German
double-quote style (Czech, German), convert punctuation to language-specific characters (Japanese, Chinese).

\begin{table*}[t]
    \centering 
    \begin{tabular}{lcccccc}
    \toprule
    \textbf{Models}        & \textbf{Zh$\rightarrow$En} & \textbf{De$\rightarrow$En} & \textbf{Cs$\rightarrow$En} & \textbf{Ru$\rightarrow$En} & \textbf{Ja$\rightarrow$En} & \textbf{$\Delta$}\\ \hline
    \textbf{Baseline}      & 16.5           & 3.5             & 40.1            & 8.5             & 21.5           &- \\
    \textbf{Vega-MT}    & \textbf{45.1}   & \textbf{58.0}   & \textbf{74.7}   & \textbf{64.9}   & 40.6           &\textbf{\textit{+38.6}}\\ \hdashline
    {Best Official} & 45.1            & 58.0            & 74.7            & 64.9            & \bf 42.0            \\ \midrule
    \textbf{Models}        & \textbf{En$\rightarrow$Zh} & \textbf{En$\rightarrow$De} & \textbf{En$\rightarrow$Cs} & \textbf{En$\rightarrow$Ru} & \textbf{En$\rightarrow$Ja} &\textbf{$\Delta$}\\\hline
    \textbf{Baseline}      & 26.6            & -40.6           & 66.9            & -1.4            & 42.1            &-\\
    \textbf{Vega-MT}    & \textbf{61.7}   & \textbf{63.2}   & {95.3}   & \textbf{69.6}   & \textbf{65.1}  &\textbf{\textit{+52.3}} \\ \hdashline
    {Best Official} & 61.7            & 63.2            & \bf 96.0            & 69.6            & 65.1            \\ 
    \bottomrule
    \end{tabular}
    \caption{\label{tab:comet} \textbf{COMET-Scores of our submissions in WMT2022 general translation task.} ``Baseline'' indicates the performance of the baseline systems. And ``Best Official'' denotes the best results of {\zct constrained systems} in each direction.}
\end{table*}

\section{Data Preparation}
\label{sec:data}
We participated in translation of all high-resource tracks and one medium-resource track, including Chinese$\leftrightarrow$English (Zh$\leftrightarrow$En), German$\leftrightarrow$English (De$\leftrightarrow$En), Czech$\leftrightarrow$English (Cs$\leftrightarrow$En), Russian$\leftrightarrow$English (Ru$\leftrightarrow$En), and Japanese$\leftrightarrow$English (Ja$\leftrightarrow$En). 

In this section, we take the En$\leftrightarrow$Zh translation as example and describe how to prepare the training data. The setting is the same for other language pairs.
We use all available parallel corpus for En$\leftrightarrow$Zh~\footnote{both parallel and monolingual corpus can be obtained from\url{https://www.statmt.org/wmt22/translation-task.html}}, including ParaCrawl v9, News Commentary v16, Wiki Titles v3, UN Parallel Corpus V1.0, CCMT Corpus, WikiMatrix and Back-translated news. For monolingual data, we randomly sample from ``News Crawl'' and ``Common Crawl''. The final corpus statistics are presented in Table~\ref{tab:data_statics}.

To improve the quality of parallel data, we further propose to filter the low-quality samples.
First, we remove the sentence pair which is predicted as wrong language with  \texttt{Fasttext}~\cite{joulin-etal-2017-bag, joulin2016fasttext}. 
Second, we replace unicode punctuation and also normalize punctuation with \texttt{mosesdecoder}. 
We also remove duplicate sentence pairs and filter out sentences with illegal characters.
For length, we remove sentences longer than 250 words and with a source/target length ratio exceeding 3.

\section{Experiments}
\label{sec:exp}
\paragraph{Settings}
 
We use the extremely large Transformer ($\mathcal{M}_\textbf{XL}$) for all tasks and Transformer-\textsc{BIG} ($\mathcal{M}_\textbf{BIG}$) for bilingual baselines.
For $\mathcal{M}_\textbf{BIG}$, we empirically adopt large batch strategy~\cite{edunov2018understanding} (\textit{i.e.} 458K tokens/batch) to optimize the performance. The learning rate warms up to $1\times10^{-7}$ for 10K steps, and then decays for 70K steps with the cosine schedule. For regularization, we tune the dropout rate from [0.1, 0.2, 0.3] based on validation performance, and apply weight decay with 0.01 and label smoothing with $\epsilon$ = 0.1. We use Adam optimizer ~\citep{kingma2015adam} to train models. We evaluate the performance on an ensemble of last 10 checkpoints to avoid stochasticity.
For the main model $\mathcal{M}_\textbf{XL}$, we adopt 1M Tokens/Batch to optimize the performance both in multilingual pretraining and bilingual finetuning. 
We set 0.1 as the label smoothing ratio, 4000 as warm-up steps, and 1e-3 as the learning rate. We optimize Vega-MT with Adam~\cite{kingma2015adam}. We use 100k updates for multi-directional pretraining, 40k updates for each specific-directional finetuning.
For evaluation, we select SacreBLEU~\cite{post-2018-call} as the metric for all tasks. \texttt{news-test2020} and \texttt{news-test2021} are selected for validation and test respectively. 

All parallel data will be used in the multi-directional PT stage, and during specific-directional FT, corresponding bilingual data augmented by bidirectional self-training are utilized.
Each sentence are jointly tokenized in to sub-word units with  SentencePiece~\cite{kudo-richardson-2018-sentencepiece}, which is trained on all concatenated multilingual parallel data for Transformer-\textsc{XL} with merge operation 120K at the pretraining stage, and during finetuning stage, is trained on corresponding bilingual data with merge operation 60K for English and 75K for other languages.
And for each baseline with Transformer-\textsc{Big}, the joint bilingual vocab size is 80K.
During pretraining, we select the sample with temperature-based method (T=5) to preserve the representation of relatively low-resource language, \textit{e.g.} Japanese.
We grid-search the beam size within the range of [3,4,5,..,8] on validation set for each translation task.
All models are trained on 32 DGX-SuperPOD A100 GPUs {\zct for about two weeks pre-training and five days fine-tuning. }

\paragraph{Main Results}
To illustrate the effectiveness of each strategy in our Vega-MT, we report the ablation results in Table~\ref{tab:ablation} on Zh$\leftrightarrow$En tasks. Clearly, directly generating the translations with the multi-directional pretrained model could obtain average +3.2 and +1.9 BLEU improvements for Zh-En and En-Zh, respectively, which is consistent with the findings of \citet{facebook2021}. We show that tuning on downstream bilingual data could further improve the translation by +1.4 BLEU points, showing the necessity of bridging the cross-lingual gap with in-domain learning during leveraging multilingual pretrain~\cite{Zan2022BridgingCG}. 
Bidirectional self-training actually contains several strategies, \textit{e.g.} back translation, distillation and data diversification, and we empirically show that such data augmentation strategy nicely complement pretraining, which is also verified by \citet{Liu2021OnTC}.
Other strategies could consistently enhance the translation performance besides the generalization FT for the news domain test2021, where the Zh-En model decreases the BLEU scores (-0.8 BLEU) because the generalization FT is designed and tuned for the general domain test2022.

Table~\ref{tab:sacrebleu} and Table~\ref{tab:comet} show the final submissions in terms of SacreBLEU and COMET scores, including Zh, De, Cs, Ru and Ja to/from En, listing the baseline and our final submissions.
We also report the best official scores among all constrained systems ``Best Official'' as reference.
As seen, SacreBLEU and COMET results show identical trends, where our Vega-MT outperforms baseline Transformer-\textsc{Big} by +11.2/ +38.6 and +14.0/ +52.3 BLEU/ COMET points, showing the effectiveness and universality of our model. Interestingly, we observe that the improvements upon En-X are more significant than that of X-En, which will be investigated in our future work.
For more system rankings, please refer Table~\ref{tab:sacrebleu_more} and Table~\ref{tab:comet_more} in Appendix for SacreBLEU and COMET results, respectively.

\section{Conclusion}
\label{sec:con}
This paper presents the JD Explore Academy machine translation system Vega-MT for WMT 2022 {\zct shared tasks on general machine translation}. We investigate various frameworks, including autoregressive and non-autoregressive Transformer with \textsc{Base}, \textsc{Big} and \textsc{XL} settings, respectively, to build strong baseline models. Then we push the limit of bidirectional training by scaling up two main factors, \textit{i.e.} language pairs and model scales, to develop the powerful foundation Vega-MT model. Also, the popular data augmentation methods, \textit{e.g.} cycle translation and bidirectional self-training, are combined to improve their performance. We carefully design the generalization adaptation strategies to further improve the multi-domain performance. 
Among all participated constrained systems, our Vega-MT won 7 champions, 2 runners-up, and 1 third place w.r.t sacreBLEU. And according to the COMET, we won 8 champions and 2 runners-up.

\section*{Acknowledgments}
This work was partially supported by the Major Science and Technology Innovation 2030 ”New Generation Artificial Intelligence” key project (No. 2021ZD0111700).
The authors wish to thank the organizers of WMT2022 for their great efforts in the organization, and their prompt responses to our questions. The authors are grateful to the anonymous reviewers for their insightful comments and careful proofreading.
The authors also specially thank Yukang Zhang (JD Explore Academy), who kindly supports us by maintaining a stable computing platform.

\appendix

\begin{table*}[t]
    \centering 
    \scalebox{0.9}{
    \begin{tabular}{cccccc}
    \toprule
    \textbf{pair}  & \textbf{system}     & \textbf{id} & \textbf{is\_constrained} & \textbf{metric} & \textbf{score} \\ \hline
    \textbf{En-Cs} & Lan-Bridge          & 551         & FALSE                   & bleu-B          & 45.6           \\
    \textbf{En-Cs} & JDExploreAcademy    & 829         & TRUE                    & bleu-B          & \textbf{41.4}  \\
    \textbf{En-Cs} & CUNI-DocTransformer & 800         & TRUE                    & bleu-B          & 39.8           \\
    \textbf{En-Cs} & CUNI-Bergamot       & 734         & TRUE                    & bleu-B          & 38.6           \\
    \textbf{En-Cs} & CUNI-Transformer    & 761         & TRUE                    & bleu-B          & 37.7           \\ \hline
    \textbf{pair}  & \textbf{system}     & \textbf{id} & \textbf{is\_constrained} & \textbf{metric} & \textbf{score} \\ \hline
    \textbf{En-De} & JDExploreAcademy    & 843         & TRUE                    & bleu-A          & \textbf{37.8}  \\
    \textbf{En-De} & Lan-Bridge          & 549         & FALSE                   & bleu-A          & 36.1           \\
    \textbf{En-De} & PROMT               & 694         & FALSE                   & bleu-A          & 36.1           \\
    \textbf{En-De} & OpenNMT             & 207         & FALSE                   & bleu-A          & 35.7           \\ \hline
    \textbf{pair}  & \textbf{system}     & \textbf{id} & \textbf{is\_constrained} & \textbf{metric} & \textbf{score} \\ \hline
    \textbf{En-Ja} & NT5                 & 763         & TRUE                    & bleu-A          & 42.5           \\
    \textbf{En-Ja} & DLUT                & 789         & TRUE                    & bleu-A          & 41.8           \\
    \textbf{En-Ja} & LanguageX           & 676         & FALSE                   & bleu-A          & 41.7           \\
    \textbf{En-Ja} & JDExploreAcademy    & 516         & TRUE                    & bleu-A          & \textbf{41.5}  \\
    \textbf{En-Ja} & Lan-Bridge          & 555         & FALSE                   & bleu-A          & 39.4           \\ \hline
    \textbf{pair}  & \textbf{system}     & \textbf{id} & \textbf{is\_constrained} & \textbf{metric} & \textbf{score} \\ \hline
    \textbf{En-Ru} & JDExploreAcademy    & 509         & TRUE                    & bleu-A          & \textbf{32.7}  \\
    \textbf{En-Ru} & Lan-Bridge          & 556         & FALSE                   & bleu-A          & 32.6           \\
    \textbf{En-Ru} & HuaweiTSC           & 680         & TRUE                    & bleu-A          & 30.8           \\
    \textbf{En-Ru} & PROMT               & 804         & FALSE                   & bleu-A          & 30.6           \\
    \textbf{En-Ru} & SRPOL               & 265         & TRUE                    & bleu-A          & 30.4           \\ \hline
    \textbf{pair}  & \textbf{system}     & \textbf{id} & \textbf{is\_constrained} & \textbf{metric} & \textbf{score} \\ \hline
    \textbf{En-Zh} & LanguageX           & 716         & FALSE                   & bleu-A          & 54.3           \\
    \textbf{En-Zh} & HuaweiTSC           & 557         & FALSE                   & bleu-A          & 49.7           \\
    \textbf{En-Zh} & JDExploreAcademy    & 834         & TRUE                    & bleu-A          & \textbf{49.7}  \\
    \textbf{En-Zh} & AISP-SJTU           & 611         & TRUE                    & bleu-A          & 48.8           \\
    \textbf{En-Zh} & Manifold            & 336         & TRUE                    & bleu-A          & 48.7           \\ \hline
    \textbf{pair}  & \textbf{system}     & \textbf{id} & \textbf{is\_constrained} & \textbf{metric} & \textbf{score} \\ \hline
    \textbf{Cs-En} & JDExploreAcademy    & 505         & TRUE                    & bleu-B          & \textbf{54.9}  \\
    \textbf{Cs-En} & Lan-Bridge          & 585         & FALSE                   & bleu-B          & 54.5           \\
    \textbf{Cs-En} & CUNI-DocTransformer & 805         & TRUE                    & bleu-B          & 51.9           \\
    \textbf{Cs-En} & CUNI-Transformer    & 772         & TRUE                    & bleu-B          & 51.6           \\
    \textbf{Cs-En} & SHOPLINE-PL         & 819         & TRUE                    & bleu-B          & 46.8           \\ \hline
    \textbf{pair}  & \textbf{system}     & \textbf{id} & \textbf{is\_constrained} & \textbf{metric} & \textbf{score} \\ \hline
    \textbf{De-En} & JDExploreAcademy    & 809         & TRUE                    & bleu-A          & \textbf{33.7}  \\
    \textbf{De-En} & Lan-Bridge          & 587         & FALSE                   & bleu-A          & 33.4           \\
    \textbf{De-En} & PROMT               & 796         & FALSE                   & bleu-A          & 32.5           \\
    \textbf{De-En} & LT22                & 605         & TRUE                    & bleu-A          & 26.0           \\ \hline
    \textbf{pair}  & \textbf{system}     & \textbf{id} & \textbf{is\_constrained} & \textbf{metric} & \textbf{score} \\ \hline
    \textbf{Ja-En} & NT5                 & 766         & TRUE                    & bleu-A          & 26.6           \\
    \textbf{Ja-En} & JDExploreAcademy    & 512         & TRUE                    & bleu-A          & \textbf{25.6}  \\
    \textbf{Ja-En} & DLUT                & 693         & TRUE                    & bleu-A          & 24.8           \\
    \textbf{Ja-En} & Lan-Bridge          & 588         & FALSE                   & bleu-A          & 22.8           \\
    \textbf{Ja-En} & NAIST-NICT-TIT      & 583         & TRUE                    & bleu-A          & 22.7           \\ \hline
    \textbf{pair}  & \textbf{system}     & \textbf{id} & \textbf{is\_constrained} & \textbf{metric} & \textbf{score} \\ \hline
    \textbf{Ru-En} & Lan-Bridge          & 589         & FALSE                   & bleu-A          & 45.2           \\
    \textbf{Ru-En} & HuaweiTSC           & 836         & TRUE                    & bleu-A          & 45.1           \\
    \textbf{Ru-En} & JDExploreAcademy    & 769         & TRUE                    & bleu-A          & \textbf{45.1}  \\
    \textbf{Ru-En} & SRPOL               & 666         & TRUE                    & bleu-A          & 43.6           \\
    \textbf{Ru-En} & ALMAnaCH-Inria      & 710         & TRUE                    & bleu-A          & 30.3           \\ \hline
    \textbf{pair}  & \textbf{system}     & \textbf{id} & \textbf{is\_constrained} & \textbf{metric} & \textbf{score} \\ \hline
    \textbf{Zh-En} & JDExploreAcademy    & 708         & TRUE                    & bleu-A          & \textbf{33.5}  \\
    \textbf{Zh-En} & LanguageX           & 219         & FALSE                   & bleu-A          & 31.9           \\
    \textbf{Zh-En} & HuaweiTSC           & 477         & FALSE                   & bleu-A          & 29.8           \\
    \textbf{Zh-En} & AISP-SJTU           & 648         & TRUE                    & bleu-A          & 29.7           \\
    \textbf{Zh-En} & Lan-Bridge          & 386         & FALSE                   & bleu-A          & 28.1           \\
    \bottomrule
    \end{tabular}}
    \caption{\label{tab:sacrebleu_more} \textbf{Ranking of our submissions in terms of SacreBLEU-Score} in WMT2022 general translation task.}
\end{table*}

\begin{table*}[t]
    \centering 
    \scalebox{0.9}{
    \begin{tabular}{cccccc}
    \toprule
    \textbf{pair}  & \textbf{system}     & \textbf{id} & \textbf{is\_constrained} & \textbf{metric} & \textbf{score} \\ \hline
    \textbf{En-Cs} & CUNI-Bergamot       & 734         & TRUE                    & COMET-B         & 0.960          \\
    \textbf{En-Cs} & JDExploreAcademy    & 829         & TRUE                    & COMET-B         & \textbf{0.953} \\
    \textbf{En-Cs} & Lan-Bridge          & 551         & FALSE                   & COMET-B         & 0.947          \\
    \textbf{En-Cs} & CUNI-DocTransformer & 800         & TRUE                    & COMET-B         & 0.917          \\
    \textbf{En-Cs} & CUNI-Transformer    & 761         & TRUE                    & COMET-B         & 0.866          \\\hline
    \textbf{pair}  & \textbf{system}     & \textbf{id} & \textbf{is\_constrained} & \textbf{metric} & \textbf{score} \\\hline
    \textbf{En-De} & JDExploreAcademy    & 843         & TRUE                    & COMET-A         & \textbf{0.632} \\
    \textbf{En-De} & Lan-Bridge          & 549         & FALSE                   & COMET-A         & 0.588          \\
    \textbf{En-De} & OpenNMT             & 207         & FALSE                   & COMET-A         & 0.572          \\
    \textbf{En-De} & PROMT               & 694         & FALSE                   & COMET-A         & 0.558          \\\hline
    \textbf{pair}  & \textbf{system}     & \textbf{id} & \textbf{is\_constrained} & \textbf{metric} & \textbf{score} \\\hline
    \textbf{En-Ja} & JDExploreAcademy    & 516         & TRUE                    & COMET-A         & \textbf{0.651} \\
    \textbf{En-Ja} & NT5                 & 763         & TRUE                    & COMET-A         & 0.641          \\
    \textbf{En-Ja} & LanguageX           & 676         & FALSE                   & COMET-A         & 0.621          \\
    \textbf{En-Ja} & DLUT                & 789         & TRUE                    & COMET-A         & 0.605          \\
    \textbf{En-Ja} & Lan-Bridge          & 555         & FALSE                   & COMET-A         & 0.565          \\\hline
    \textbf{pair}  & \textbf{system}     & \textbf{id} & \textbf{is\_constrained} & \textbf{metric} & \textbf{score} \\\hline
    \textbf{En-Ru} & JDExploreAcademy    & 509         & TRUE                    & COMET-A         & \textbf{0.696} \\
    \textbf{En-Ru} & Lan-Bridge          & 556         & FALSE                   & COMET-A         & 0.673          \\
    \textbf{En-Ru} & PROMT               & 804         & FALSE                   & COMET-A         & 0.603          \\
    \textbf{En-Ru} & SRPOL               & 265         & TRUE                    & COMET-A         & 0.597          \\
    \textbf{En-Ru} & HuaweiTSC           & 680         & TRUE                    & COMET-A         & 0.592          \\\hline
    \textbf{pair}  & \textbf{system}     & \textbf{id} & \textbf{is\_constrained} & \textbf{metric} & \textbf{score} \\\hline
    \textbf{En-Zh} & LanguageX           & 716         & FALSE                   & COMET-A         & 0.638          \\
    \textbf{En-Zh} & JDExploreAcademy    & 834         & TRUE                    & COMET-A         & \textbf{0.617} \\
    \textbf{En-Zh} & Lan-Bridge          & 714         & FALSE                   & COMET-A         & 0.614          \\
    \textbf{En-Zh} & Manifold            & 336         & TRUE                    & COMET-A         & 0.601          \\
    \textbf{En-Zh} & HuaweiTSC           & 557         & FALSE                   & COMET-A         & 0.595          \\\hline
    \textbf{pair}  & \textbf{system}     & \textbf{id} & \textbf{is\_constrained} & \textbf{metric} & \textbf{score} \\\hline
    \textbf{Cs-En} & JDExploreAcademy    & 505         & TRUE                    & COMET-B         & \textbf{0.747} \\
    \textbf{Cs-En} & Lan-Bridge          & 585         & FALSE                   & COMET-B         & 0.718          \\
    \textbf{Cs-En} & CUNI-DocTransformer & 805         & TRUE                    & COMET-B         & 0.706          \\
    \textbf{Cs-En} & CUNI-Transformer    & 772         & TRUE                    & COMET-B         & 0.692          \\
    \textbf{Cs-En} & SHOPLINE-PL         & 819         & TRUE                    & COMET-B         & 0.611          \\\hline
    \textbf{pair}  & \textbf{system}     & \textbf{id} & \textbf{is\_constrained} & \textbf{metric} & \textbf{score} \\\hline
    \textbf{De-En} & JDExploreAcademy    & 809         & TRUE                    & COMET-A         & \textbf{0.580} \\
    \textbf{De-En} & Lan-Bridge          & 587         & FALSE                   & COMET-A         & 0.565          \\
    \textbf{De-En} & PROMT               & 796         & FALSE                   & COMET-A         & 0.518          \\
    \textbf{De-En} & LT22                & 605         & TRUE                    & COMET-A         & 0.256          \\\hline
    \textbf{pair}  & \textbf{system}     & \textbf{id} & \textbf{is\_constrained} & \textbf{metric} & \textbf{score} \\\hline
    \textbf{Ja-En} & NT5                 & 766         & TRUE                    & COMET-A         & 0.420          \\
    \textbf{Ja-En} & JDExploreAcademy    & 512         & TRUE                    & COMET-A         & \textbf{0.406} \\
    \textbf{Ja-En} & DLUT                & 693         & TRUE                    & COMET-A         & 0.372          \\
    \textbf{Ja-En} & NAIST-NICT-TIT      & 583         & TRUE                    & COMET-A         & 0.334          \\
    \textbf{Ja-En} & LanguageX           & 435         & FALSE                   & COMET-A         & 0.329          \\\hline
    \textbf{pair}  & \textbf{system}     & \textbf{id} & \textbf{is\_constrained} & \textbf{metric} & \textbf{score} \\\hline
    \textbf{Ru-En} & JDExploreAcademy    & 769         & TRUE                    & COMET-A         & \textbf{0.649} \\
    \textbf{Ru-En} & Lan-Bridge          & 589         & FALSE                   & COMET-A         & 0.631          \\
    \textbf{Ru-En} & HuaweiTSC           & 836         & TRUE                    & COMET-A         & 0.609          \\
    \textbf{Ru-En} & SRPOL               & 666         & TRUE                    & COMET-A         & 0.595          \\
    \textbf{Ru-En} & ALMAnaCH-Inria      & 710         & TRUE                    & COMET-A         & 0.268          \\\hline
    \textbf{pair}  & \textbf{system}     & \textbf{id} & \textbf{is\_constrained} & \textbf{metric} & \textbf{score} \\\hline
    \textbf{Zh-En} & JDExploreAcademy    & 708         & TRUE                    & COMET-A         & \textbf{0.451} \\
    \textbf{Zh-En} & LanguageX           & 219         & FALSE                   & COMET-A         & 0.449          \\
    \textbf{Zh-En} & Lan-Bridge          & 386         & FALSE                   & COMET-A         & 0.430          \\
    \textbf{Zh-En} & HuaweiTSC           & 477         & FALSE                   & COMET-A         & 0.428          \\
    \textbf{Zh-En} & AISP-SJTU           & 648         & TRUE                    & COMET-A         & 0.416          \\
    \bottomrule
    \end{tabular}}
    \caption{\label{tab:comet_more} \textbf{Ranking of our submissions in terms of COMET-Score} in WMT2022 general translation task.}
\end{table*}



\bibliography{wmt2022}
\bibliographystyle{acl_natbib}
\end{document}